\documentclass[runningheads]{llncs}
\usepackage{lettrine}
\usepackage{cite}
\usepackage{graphicx}
\graphicspath{{./figures/}}
\usepackage[labelformat=parens, labelsep=space]{subcaption}
\usepackage{lipsum} 
\usepackage[T1]{fontenc}
\usepackage[utf8]{inputenc}
\usepackage{mathtools} 

\DeclarePairedDelimiter\floor{\lfloor}{\rfloor}
\newcommand{\bigdelta}{\raisebox{-.35\baselineskip}{\huge\ensuremath{\Omega}}}
\usepackage[linesnumbered,algoruled,boxed,lined]{algorithm2e}
\SetKwInOut{Input}{input}
\SetKwInOut{Output}{output}
\SetKw{Continue}{continue}
\SetKw{Break}{break}

\usepackage{textcomp}
\usepackage[table, xcdraw]{xcolor} 
\usepackage{diagbox}
\usepackage{multirow,booktabs,color,soul,threeparttable}
\definecolor{hl}{RGB}{255,255,0}
\sethlcolor{hl}
\DeclareCaptionFont{tablecaptionfont}{\fontsize{9}{9.6}\sc}
\usepackage[colorlinks,citecolor=red,urlcolor=blue,bookmarks=false,hypertexnames=true]{hyperref} 

\begin{document}
\title{Niching Diversity Estimation for Multi-modal Multi-objective Optimization}
\author{Yiming Peng, Hisao Ishibuchi\thanks{Corresponding author: Hisao Ishibuchi, hisao@sustech.edu.cn.}}
\institute{Guangdong Provincial Key Laboratory of Brain-inspired Intelligent Computation, Department of Computer Science and Engineering, \\Southern University of Science and Technology, Shenzhen 518055, China.\\
    \email{11510035@mail.sustech.edu.cn, hisao@sustech.edu.cn}}
\maketitle              
\begin{abstract}
    Niching is an important and widely used technique in evolutionary multi-objective optimization. Its applications mainly focus on maintaining diversity and avoiding early convergence to local optimum. Recently, a special class of multi-objective optimization problems, namely, multi-modal multi-objective optimization problems (MMOPs), started to receive increasing attention. In MMOPs, a solution in the objective space may have multiple inverse images in the decision space, which are termed as equivalent solutions. Since equivalent solutions are overlapping (i.e., occupying the same position) in the objective space, standard diversity estimators such as crowding distance are likely to select one of them and discard the others, which may cause diversity loss in the decision space. In this study, a general niching mechanism is proposed to make standard diversity estimators more efficient when handling MMOPs. In our experiments, we integrate our proposed niching diversity estimation method into SPEA2 and NSGA-II and evaluate their performance on several MMOPs. Experimental results show that the proposed niching mechanism notably enhances the performance of SPEA2 and NSGA-II on various MMOPs.

\end{abstract}
\section{Introduction}
\label{sec: Introduction}
\lettrine[findent=2pt]{\textbf{M}}{ }ulti-objective optimization problems (MOPs), which require an optimizer to optimize multiple conflicting objective functions simultaneously, have been actively studied in the past few decades. For consistency, in this paper, all objective functions are assumed to be converted to minimization problems. Due to the trade-off between conflicting objective functions, most MOPs have a set of Pareto optimal solutions (i.e., Pareto set), which cannot be dominated by any solutions. The projection of the Pareto set in the objective space is called the Pareto front. Generally, multi-objective optimization algorithms (MOEAs) try to find a solution set with good convergence (i.e., close to the Pareto front) and good diversity (i.e., well-distributed over the Pareto front). Therefore, diversity maintenance is a critical research topic across the field of evolutionary multi-objective optimization.

Most MOEAs are equipped with some diversity maintenance mechanisms. Naturally, diversity estimation, which is a procedure of assigning a numerical value to each solution reflecting its diversity, is a prerequisite for diversity maintenance. In the past few decades, various of diversity estimation methods have been developed along with the development of MOEAs. For instance, the well-known Pareto-based MOEA called NSGA-II \cite{NSGAII} selects solutions based on their Pareto ranks (as a primary criterion) obtained from the non-dominated sorting procedure and their crowding distance values (as a secondary criterion). In NSGA-II, crowding distance is a diversity estimator to estimate the diversity of each solution for environmental selection. Another Pareto-based algorithm called SPEA2 \cite{SPEA2} uses the Euclidean distance from each solution to its $k$-th nearest neighbor in the objective space to estimate the diversity of that solution. In the grid-based MOEA named PESA-II \cite{PESAII}, the diversity of each solution is given by the number of the solutions located in the same hyperbox in the objective space. As pointed out in \cite{SDE}, although the approaches used in different diversity estimators vary, the main idea is to estimate the diversity for a solution by measuring the similarity degree between that solution and other solutions in the population. In this paper, the term "diversity" always refers to the diversity in the objective space if not specified. Notice that larger diversity values are more preferable than smaller values.

Recently, multi-modal multi-objective optimization has become an active research topic. In MMOPs, the mapping from the decision space to the objective space is a many-to-one mapping instead of one-to-one mappings in standard MOPs. That is, multiple solutions with different decision values can have the same objective values. Such kind of solutions are termed as equivalent solutions. Fig. \ref{fig: MMOP example} gives an example of MMOP where solutions marked with the same number have the same objective values. Since it is unlikely for a real-world multi-objective optimization problem to have multiple solutions with exactly the same objective values, the definition of equivalent solutions can be relaxed as follows \cite{tanabe2020review}: 

\begin{figure}[htbp]
	\centering
	\includegraphics[width=.75\textwidth]{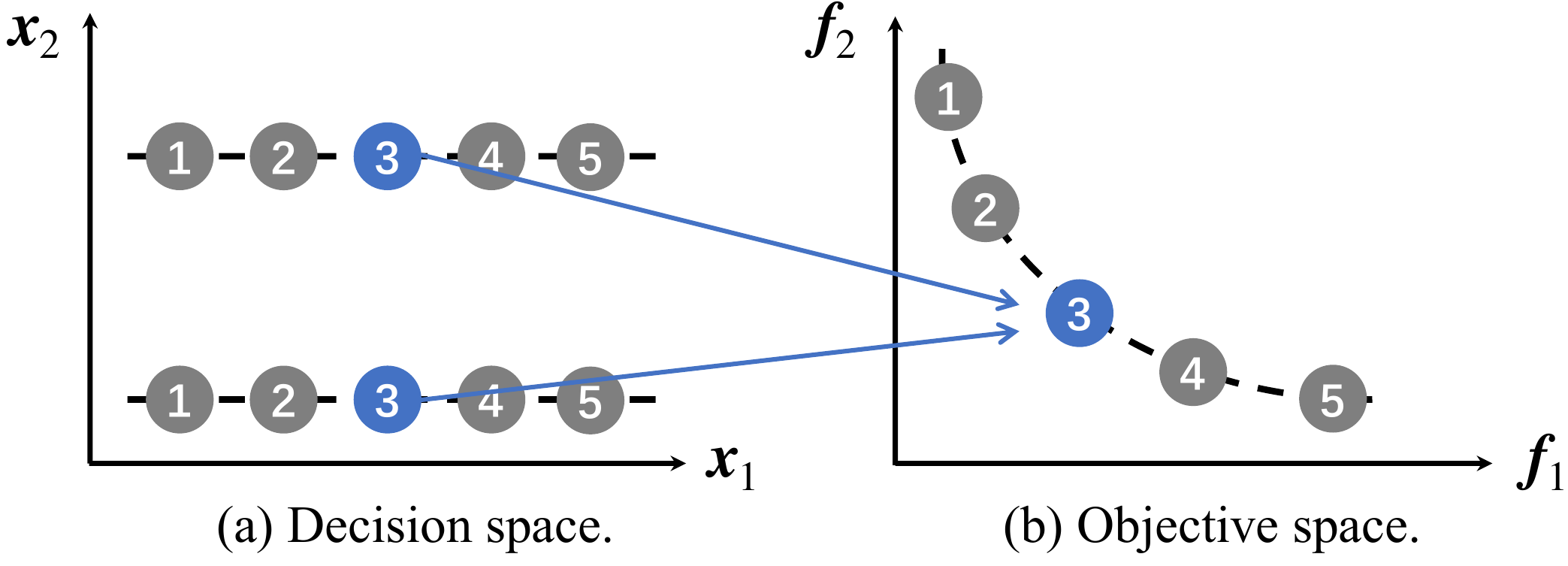}
	\caption{Illustration of MMOP. The left and right figures show the decision and objective spaces, respectively, and the dash lines in (a) and (b) denote the Pareto set and Pareto front, respectively. Circles marked with the same number are solutions that have the same objective values.}
	\label{fig: MMOP example}
\end{figure}

\begin{definition}[Equivalent Solutions]
	Solutions $\boldsymbol{x}_1$ and $\boldsymbol{x}_2$ are equivalent solutions iff $d(\boldsymbol{F}(\boldsymbol{x}_1),\boldsymbol{F}(\boldsymbol{x}_2)) \leq \delta$,
	\label{def: equivalent solutions}
\end{definition}
where $\boldsymbol{F}$ is a vector containing all objective functions, $d$ denotes the Euclidean distance function, and $\delta$ is a positive threshold parameter specified by the end user.

Solving MMOPs is meaningful since they are common in many real-world applications, e.g., rocket engine design problems \cite{MMOP_rocket} and the space mission design problems \cite{space-mission-design} both can be formulated as MMOPs. In some engineering problems, obtained solutions can become infeasible or difficult to implement due to dynamically changing environments and constraints\cite{MMOEADC}. In this regard, equivalent solutions will be able to provide alternative implementations for the decision maker. For this reason, when solving MMOPs, it is a good strategy to try to search for as many equivalent solutions as possible if no user preference is given.

As pointed out in the literature \cite{MOEAD_AD}, standard MOEAs are usually unable to preserve multiple equivalent solutions. Since equivalent solutions are located in the same (or almost the same) position(s) in the objective space, diversity estimators tend to assign high density values to all of them. As a result, equivalent solutions are usually not preferable and likely to be removed in the environmental selection procedure, which leads to the failure of solving MMOPs. To tackle this problem, we propose the use of a simple niching strategy to make standard diversity estimators more eﬃcient when handling MMOPs. Our approach is a simple, efficient, and parameterless mechanism which can be integrated into general diversity estimators in existing MOEAs.

The rest of the paper is organized as follows. Section \ref{sec: Related Work} revisits some representative diversity estimators in MOEAs and discusses their difficulties in the handling of MMOPs. In addition, Section \ref{sec: Related Work} also introduces some existing approaches for multi-modal multi-objective optimization. Next, Section \ref{sec: Proposed method} outlines our proposed niching diversity estimation method. Section \ref{sec: Experiments} reports experimental results. Lastly, concluding remarks and suggested future research directions are presented in Section \ref{sec: Conclusion}.
\section{Related Work}
\label{sec: Related Work}
In this section, we introduce two representative diversity estimators and discuss the difficulties they meet when handling MMOPs. Subsequently, some existing multi-modal multi-objective optimization algorithms are reviewed.
\subsection{Review of diversity estimators}
\subsubsection{Density in SPEA2}
In SPEA2 \cite{SPEA2}, each solution is assigned a density value which is used to calculate its fitness value. Eq. (\ref{eq: Density in SPEA2}) gives the density of a solution $\boldsymbol{x}$.
\begin{equation}
	\textit{Density} (\boldsymbol{x}) = \frac{1}{\sigma_k(\boldsymbol{x}) + 2},
	\label{eq: Density in SPEA2}
\end{equation}
where $\sigma_k(\boldsymbol{x})$ is the distance from $\boldsymbol{x}$ to its $k$-th nearest neighbor in the objective space. In SPEA2, $k$ is set to the square root of the total number of solutions in the current population as a general parameter setting.

Notice that in SPEA2, higher density means worse diversity in the objective space.

\subsubsection{Crowding distance}
Crowding distance is proposed along with the NSGA-II algorithm\cite{NSGAII} to preserve the diversity of the population in the objective space. The crowding distance of a solution $\boldsymbol{x}$ is given by the average side length of the hypercube constructed by its left and right neighbors in each objective. More precisely, for each objective, the left and right neighbors of $\boldsymbol{x}$ are the solutions at the left and right positions of $\boldsymbol{x}$ for that objective (i.e., in the list obtained by sorting the population in an increasing order of the objective values of that objective). The crowding distance of all boundary solutions (i.e., best solutions in any objectives) are set to $\infty$ to ensure that they are always selected. In NSGA-II, larger crowding distance values indicate better diversity. Formally, Eq. (\ref{eq: crowding distance}) calculates the crowding distance for a solution $\boldsymbol{x}$.
\begin{equation}
	\textit{Crowding-Distance} (\boldsymbol{x}) =
	\begin{cases}
		\infty                                                                     & ,\boldsymbol{x} \text{ is a boundary solution} \\
		\frac{1}{M}\sum_{m=1}^M[f_m(\boldsymbol{x}_{rm})-f_m(\boldsymbol{x}_{lm})] & ,\text{otherwise}
	\end{cases},
	\label{eq: crowding distance}
\end{equation}
where $M$ refers to the number of objectives, and $\boldsymbol{x}_{lm}$ and $\boldsymbol{x}_{rm}$ are the left and right neighbors of solution $\boldsymbol{x}$ regarding the $m$-th objective, respectively.
\subsection{Difficulties when handling MMOPs}
\begin{figure}
	\centering
	\includegraphics[width=.75\textwidth]{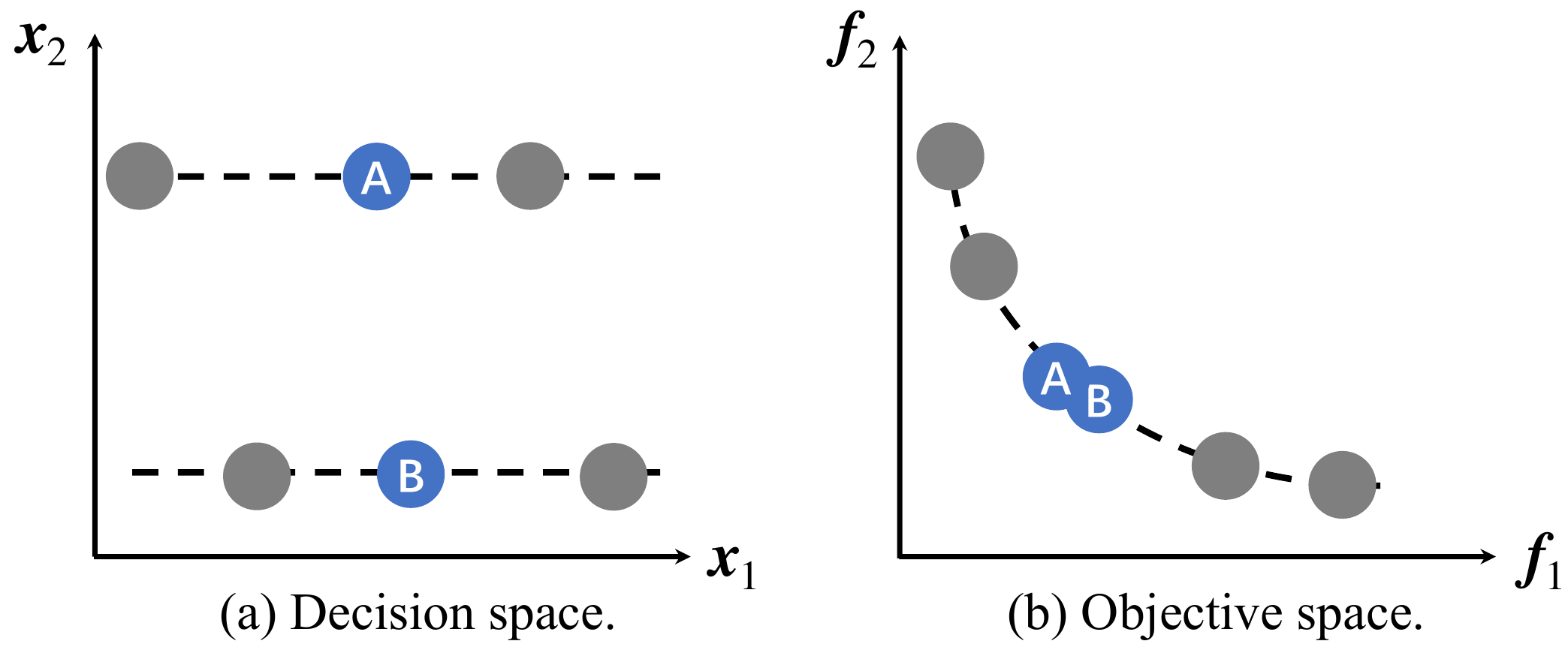}
	\caption{Explanation of the diversity loss in the decision space caused by diversity estimators when handling an MMOP. The dash lines in (a) and (b) denote the Pareto set and Pareto front, respectively.}
	\label{fig: Difficulty when handling MMOPs}
\end{figure}

In most diversity estimators in MOEAs, the solution distribution in the decision space is out of consideration, which makes them inefficient on MMOPs. As we have discussed in Section \ref{sec: Introduction}, in MMOPs, equivalent solutions have the same or almost the same objective values. Consequently, they are usually not preferable in terms of diversity (in the objective space). For this reason, diversity estimators used in MOEAs are often responsible for the loss of equivalent solutions when tackling MMOPs. Fig. \ref{fig: Difficulty when handling MMOPs} gives an example when a diversity estimator such as crowding distance produces undesirable effects. In Fig. \ref{fig: Difficulty when handling MMOPs}, $A$ and $B$ are two Pareto optimal solutions on different (but equivalent) Pareto subsets (i.e., the upper and lower dash lines in (a)). Although $A$ and $B$ have similar objective values, the decision maker may want to keep both of them since they represent different implementations (i.e., they are different in the decision space). However, a diversity estimator tends to assign bad diversity values to them due to the small difference between their objective values. As a result, some of them are likely to be removed. From this example, we can see that solutions in different regions in the decision space should be considered separately when estimating solution diversity for MMOPs. Following this idea, we propose a niching diversity estimation method in Section \ref{sec: Proposed method}.

\subsection{Multi-modal multi-objective optimization algorithms}
\label{sec: Existing multi-modal multi-objective optimization algorithms}
In most state-of-the-art multi-modal multi-objective evolutionary algorithms (MMEAs), the diversity in the decision space is maintained by niching strategies. Some MMEAs extend existing niching strategies in MOEAs to enable them to maintain the diversity in the objective space as well as in the decision space. For example, in \cite{OmniOptimizer}, Deb and Tiwari proposed one of the first MMEA called Omni-optimizer which modifies the crowding distance to measure the diversity in the decision space and the objective space simultaneously. Yue et. al. proposed a particle swarm optimizer named MO\_Ring\_PSO\_SCD \cite{MO_Ring_PSO_SCD} which adopts a similar modified crowding distance and a ring topology to create a niche structure. The DNEA algorithm \cite{DNEA} applies the fitness sharing \cite{Sharing} to both decision and objective spaces and combines them into a single sharing function. Some MMEAs are proposed with dedicated niching strategies in the decision space. Tanabe et. al. proposed a decomposition-based MMEA called MOEA/D-AD\cite{MOEAD_AD} where multiple solutions can be assigned to a weight vector, and a newly generated solution only competes with other solutions which are assigned to the same weight vector and neighboring to that solution in the decision space. In our previous study \cite{MOEAD_MM}, we proposed another decomposition-based MMEA which utilizes a clearing strategy in the decision space. Some MMEAs such as the algorithms proposed in \cite{DBSCAN_MMEA} and \cite{MMOEADC} use clustering approaches to maintain the niching structure in the decision space.
\section{Proposed Method}
\label{sec: Proposed method}
\subsection{Niching diversity estimation}
In this section, we outline our proposed niching diversity estimation method for multi-modal multi-objective optimization. Here we first introduce a general representation for most diversity estimators used in MOEAs before diving into the details of the proposed method.

Generally, for a solution $\boldsymbol{x}_i$ in a solution set $\boldsymbol{S}$, its diversity regarding $\boldsymbol{S}$ can be expressed as follows:
\begin{equation}
	\textit{Diversity}(\boldsymbol{x}_i, \boldsymbol{S}) = \mathop{\bigdelta}_{\boldsymbol{x}_j\in\boldsymbol{S}, j \neq i}{C(\boldsymbol{x}_i, \boldsymbol{x}_j)},
	\label{eq: Standard diversity estimator}
\end{equation}
where $C$ is a function which calculates the diversity contribution from a pair of solutions $\boldsymbol{x}_i$ and $\boldsymbol{x}_j$ regarding their objective values, and $\Omega$ is an aggregation function (e.g., sum or mean) to combine the diversity contribution from each pair.
The idea of our proposed method is straightforward: to restrict the diversity estimation within a niche. We make the following simple modifications to Eq. (\ref{eq: Standard diversity estimator}):
\begin{equation}
	\textit{Niching-Diversity}(\boldsymbol{x}_i, \boldsymbol{S}) = \textit{Diversity}(\boldsymbol{x}_i, \boldsymbol{S}^\prime),
	\label{eq: Niching diversity estimator}
\end{equation}
where $\boldsymbol{S}^\prime$ contains all solutions in $\boldsymbol{S}$ which are in the same niche as $\boldsymbol{x}_i$. In our paper, the closest $k$ solutions in $\boldsymbol{S}$ to $\boldsymbol{x}_i$ in the \textbf{decision space} are considered as a niche.

From Eq. (\ref{eq: Niching diversity estimator}), we can see that the diversity estimation for each solution is limited to its neighbors in the decision space. With the niching strategy, solution distribution in the decision space is taken into consideration. Take Fig. \ref{fig: Difficulty when handling MMOPs} as an example, if $k = 2$ and crowding distance is used, the two nearest neighbors are selected for each solution (e.g., solution $A$) in the decision space, and the crowding distance is calculated using the selected neighbors in the objective space. Solution $B$ is unlikely to be chosen as a neighbor of solution $A$ (i.e., $B$ is not likely to be used for the crowding distance calculation of $A$). In this manner, the diversity of $A$ and $B$ can be estimated in a desirable manner for maintaining the decision space diversity. From this example, we can see that the proposed niching strategy can help diversity estimators in MOEAs to handle MMOPs properly with an appropriate value of $k$.

Compared to existing approaches we have discussed in Section \ref{sec: Existing multi-modal multi-objective optimization algorithms}, our proposed method does not rely on the actual implementations of diversity estimators. It is a general niching strategy that can be conveniently integrated into most diversity estimators in MOEAs.

\subsection{SPEA2 and NSGA-II with niching diversity estimation}
In this section, we select two classical MOEAs: SPEA2 and NSGA-II to demonstrate the procedure of our proposed niching diversity estimation method into MOEAs. The resulting algorithms are termed Niching-SPEA2 and Niching-NSGA-II, respectively.

In SPEA2 and NSGA-II, diversity estimation is only involved in the environmental selection procedure although the estimated diversity values may be used in other procedures. Therefore, we only describe the modified versions of environmental selection.

In the environmental selection procedure of Niching-SPEA2, the niching strategy is applied to both fitness calculation and archive truncation as outlined in lines 4 and 13 in Algorithm \ref{algo: Niching SPEA2}. In these two procedures, distance calculation is restricted by the niching strategy. For Niching-NSGA-II, in each generation, non-dominated sorting is employed to rank the whole population into several fronts. Afterward, the crowding distance is computed in each front with the proposed niching strategy.

\begin{algorithm}
	\caption{Environmental Selection Procedure of Niching-SPEA2.}
	\label{algo: Niching SPEA2}
	\Input{$\boldsymbol{P}$: input population;\\$N$: the number of survivors;}
	\Output{$\boldsymbol{Q}$: population for the next generation;}
	\tcc{Fitness assignment}
	$\boldsymbol{S} \gets$ the strength value for each solution in $\boldsymbol{P}$\;
	$\boldsymbol{R} \gets$ the raw fitness value for each solution in $\boldsymbol{P}$\;
	$\boldsymbol{D} \gets$ the \textbf{niching density value} for each solution in $\boldsymbol{P}$\;
	\For{$i=1,2,\ldots,|\boldsymbol{P}|$}{
		$\boldsymbol{F}(i) = \boldsymbol{R}(i) + \boldsymbol{D}(i)$; \tcp{fitness value}
	}
	\tcc{Archive truncation}
	$\boldsymbol{Q} \gets $ non-dominated solutions in $\boldsymbol{P}$\;
	\uIf{$|\boldsymbol{Q}| < N$}{
		$\boldsymbol{Q} \gets $ best $N$ solutions in $\boldsymbol{P}$ regarding their fitness values.
	}\Else{
		\While{$|\boldsymbol{Q}| > N$}{
			Repeatedly remove the solution with shortest distance to other solutions \textbf{in the same niche} from $\boldsymbol{Q}$.
		}
	}
\end{algorithm}

\section{Numerical Experiments}
\label{sec: Experiments}
\subsection{Experimental settings}
In our experiments, the performance of Niching-SPEA2 and Niching-NSGA-II as well as the corresponding original algorithms is evaluated on ten MMOPs. Specifically, we use the SYM-PART\cite{SYMPART}, the Omni-test\cite{OmniOptimizer}, and the MMF1--8\cite{MO_Ring_PSO_SCD} test problems for benchmarking. For the Omni-test problem, we set the number of decision variables to $3$. For the rest of test problems, the default parameter settings in the corresponding papers \cite{SYMPART, MO_Ring_PSO_SCD} are used. Each algorithm is evaluated on each test problem 31 times independently with population size 100 and 50,000 function evaluations. The niching parameter $k$ in Eq. (\ref{eq: Niching diversity estimator}) is set to $\floor{\sqrt{N}}$, where $N$ is the size of $\boldsymbol{S}$. This setting of $k$ is based on the suggestions in \cite{silverman1986density} for statistics and data analysis.

We choose two widely used indicators: $\operatorname{IGD}^+$\cite{IGDPlus} and IGDX\cite{IGDX} to evaluate the performance of an algorithm in the objective space and the decision space, respectively. Smaller $\operatorname{IGD}^+$ and IGDX values indicate better proximity of the obtained solution set to the Pareto front and the Pareto set, respectively.

\subsection{Experimental results}
\begin{figure}
	\centering
	\begin{subfigure}[b]{.49\textwidth}
		\includegraphics[width=\textwidth]{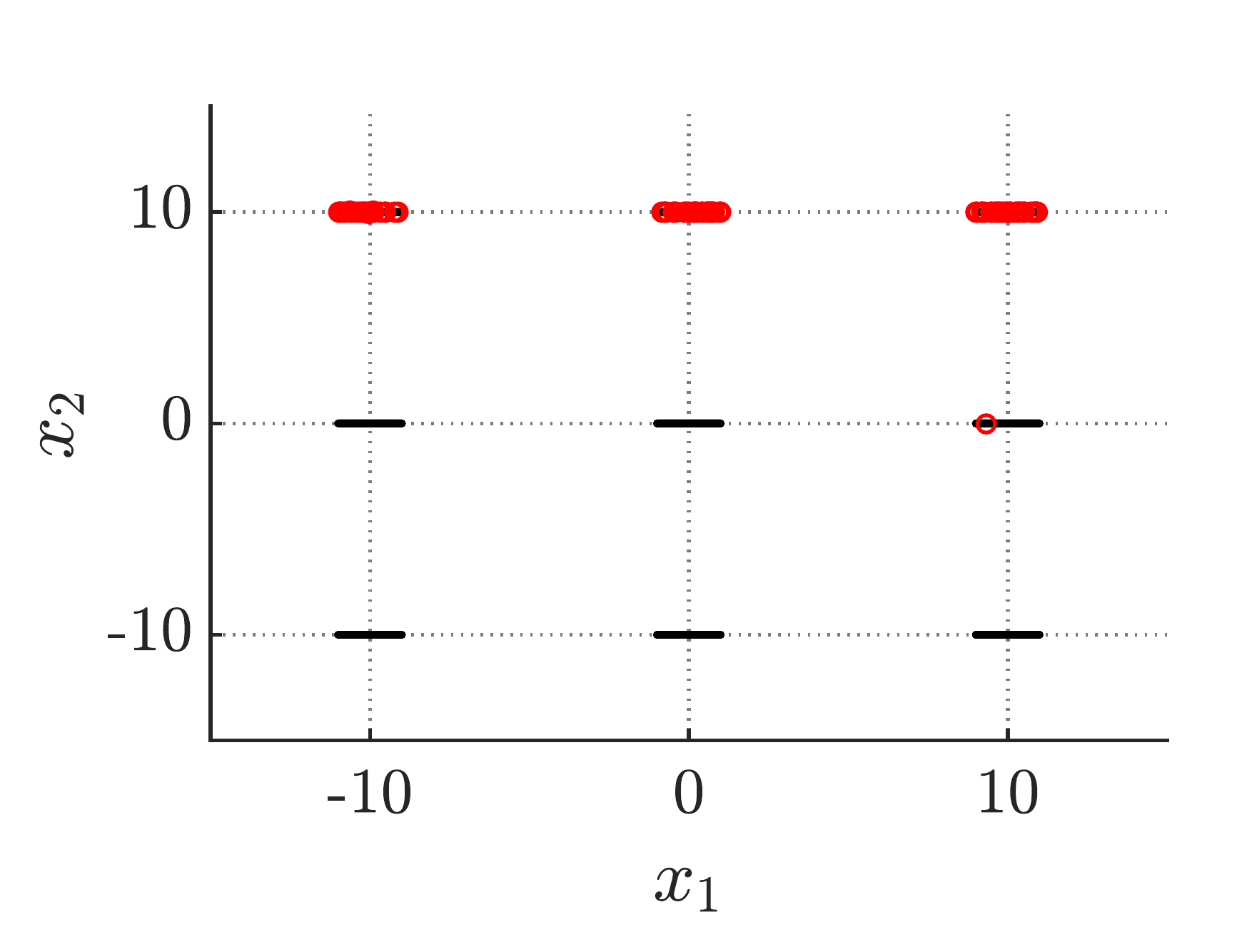}
		\caption{Decision space.}
	\end{subfigure}
	\begin{subfigure}[b]{.49\textwidth}
		\includegraphics[width=\textwidth]{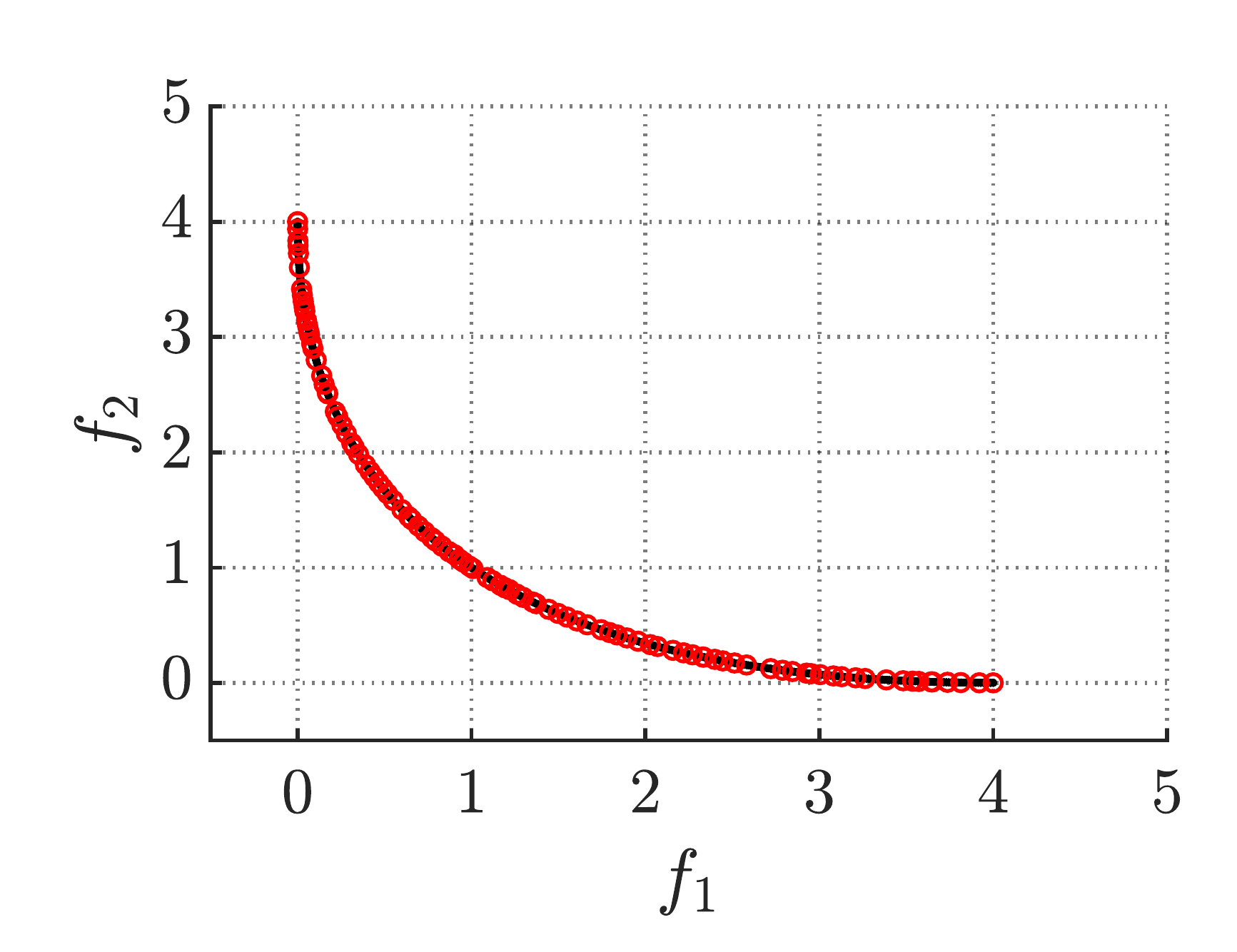}
		\caption{Objective space.}
	\end{subfigure}
	\caption{Non-dominated solutions obtained by NSGA-II on SYM-PART. The black lines show Pareto optimal solutions, and red circles show the obtained solutions.}
	\label{fig: NSGA-II on SYM-PART}
\end{figure}

\begin{figure}
	\centering
	\begin{subfigure}[b]{.49\textwidth}
		\includegraphics[width=\textwidth]{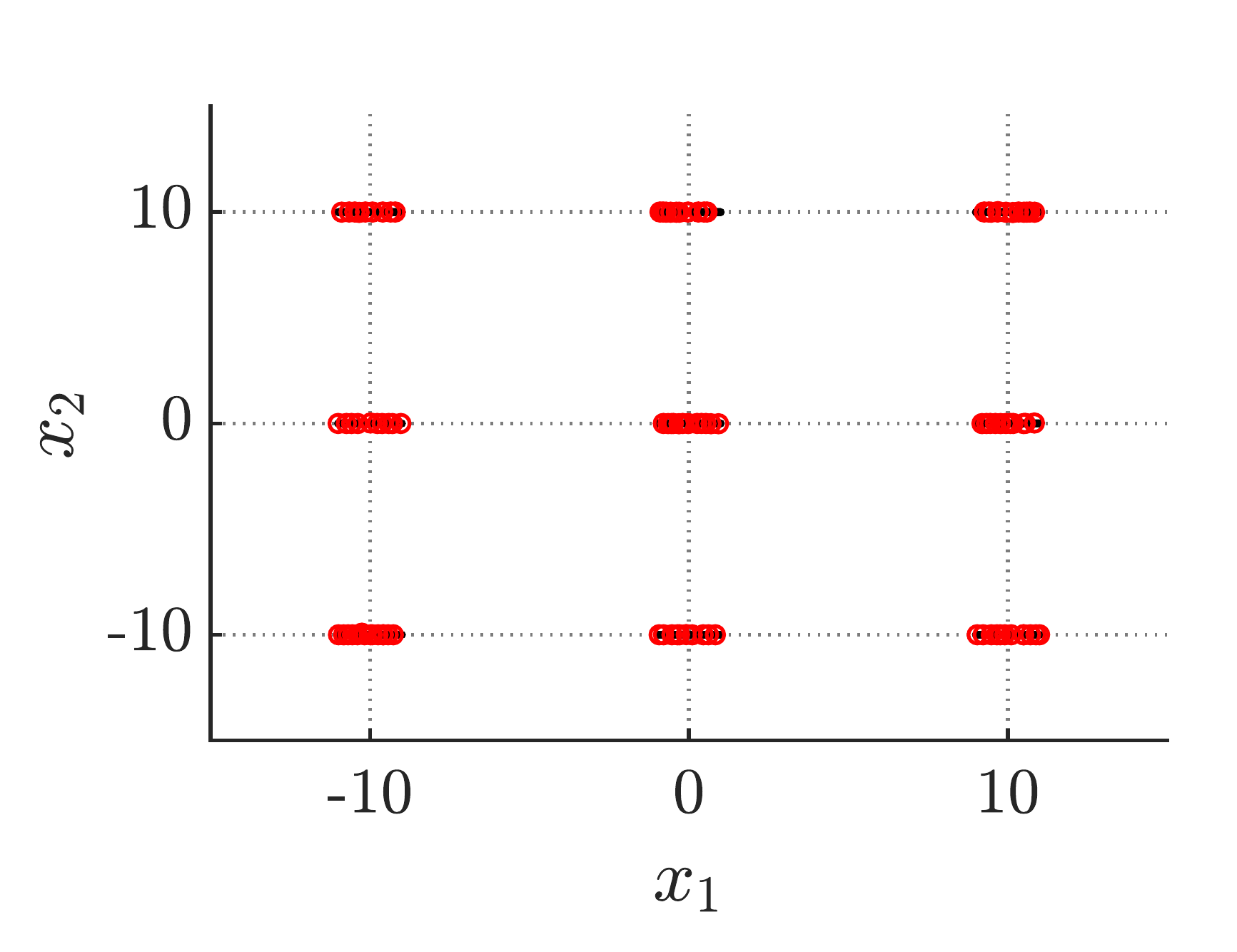}
		\caption{Decision space.}
	\end{subfigure}
	\begin{subfigure}[b]{.49\textwidth}
		\includegraphics[width=\textwidth]{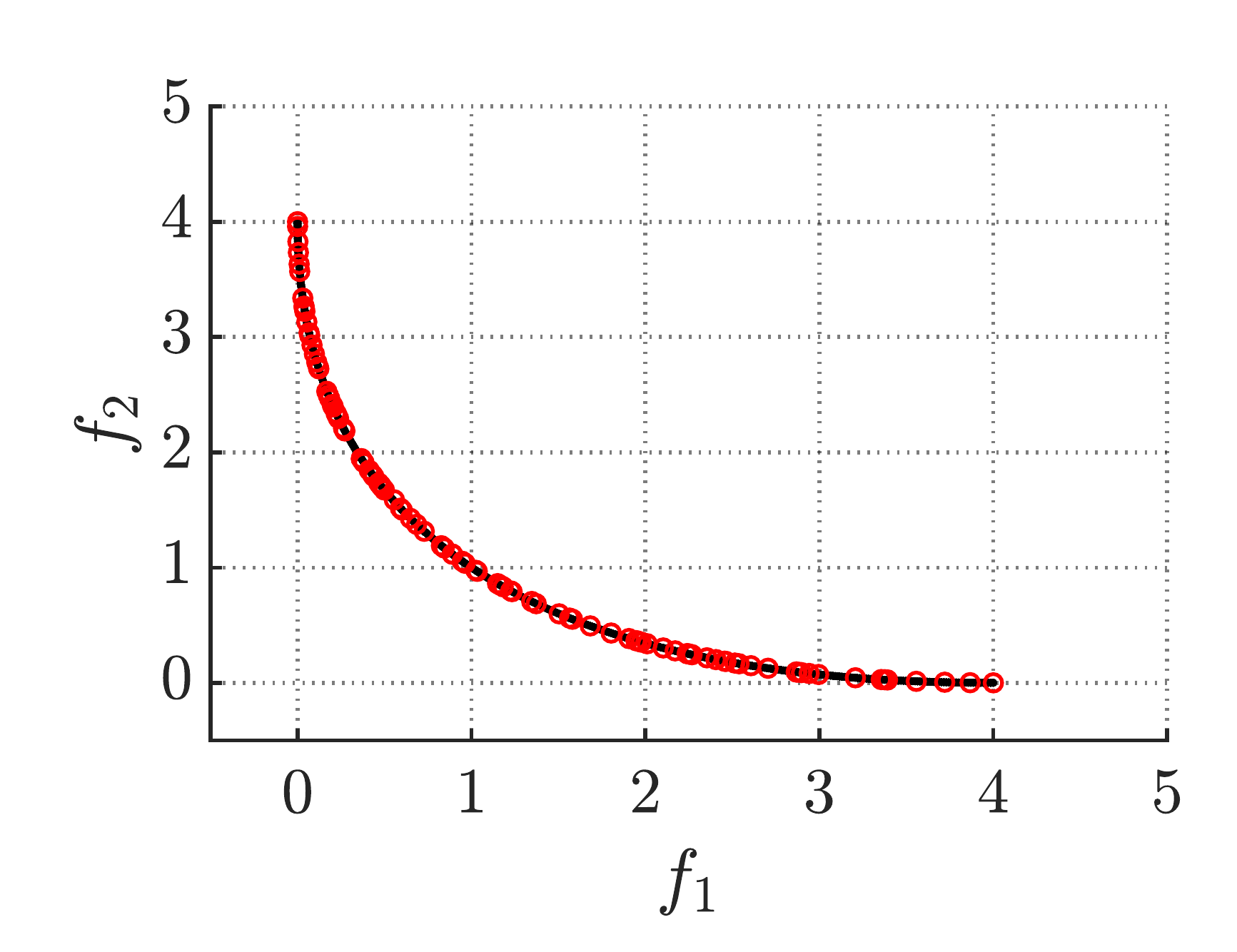}
		\caption{Objective space.}
	\end{subfigure}
	\caption{Obtained non-dominated solutions by Niching-NSGA-II on SYM-PART. The black lines show Pareto optimal solutions, and red circles show the obtained solutions.}
	\label{fig: Niching-NSGA-II on SYM-PART}
\end{figure}

To demonstrate the efficacy of our proposed niching strategy, first we visually examine the distribution of the solution sets found by Niching-NSGA-II and its original version on the SYM-PART test problem. Non-dominated solutions obtained from a single run of each algorithm are shown in Fig. \ref{fig: NSGA-II on SYM-PART} and Fig. \ref{fig: Niching-NSGA-II on SYM-PART}. In each figure, we select the run with the median $\operatorname{IGD}^+$ value among all 31 independent runs as a representative for visual examination.

Fig. \ref{fig: NSGA-II on SYM-PART} (a) clearly shows that NSGA-II is poorly performed on the SYM-PART test problem. Most obtained solutions are distributed in the upper three Pareto subsets, while almost no solution lies on the other six Pareto subsets. This is because NSGA-II is a standard MOEA without diversity maintenance mechanisms in the decision space. In comparison, Niching-NSGA-II clearly outperforms NSGA-II as shown in Fig. \ref{fig: Niching-NSGA-II on SYM-PART} (a), where all nine Pareto subsets are covered. This experimental result verifies that our proposed niching strategy can efficiently prevent the loss of equivalent solutions and preserve the diversity in the decision space. Regarding to the distribution in the objective space, in Figs. \ref{fig: NSGA-II on SYM-PART} (b) and \ref{fig: Niching-NSGA-II on SYM-PART} (b), Niching-NSGA-II slightly underperforms NSGA-II on the SYM-PART test problem. As reported in \cite{tanabe2020review}, this is because equivalent solutions have small (or even zero) contribution to the diversity in the objective space. These observations suggest that there is a clear trade-off between the diversity on the Pareto set in the decision space and the diversity on the Pareto front in the objective space when solving MMOPs.

\begin{table}
	\caption{Statistical comparison results regarding the IGDX indicator. Mean and standard deviation of IGDX values are shown. Better results are highlighted.}
	\label{table: IGDX comparison}
	\centering
	\centerline{
		\begin{tabular}{@{\hspace{.5em}}c@{\hspace{2em}}c@{\hspace{1em}}cc@{\hspace{2em}}c@{\hspace{1em}}c@{\hspace{.5em}}}
			\toprule
			                  & NSGA-II           & Niching-NSGA-II                               &  & SPEA2             & Niching-SPEA2                                       \\ \midrule
			Omni-test         & 1.2610 $\pm$ 0.35 & \cellcolor[HTML]{C0C0C0}0.3647 $\pm$ 0.15 $+$ &  & 1.3127 $\pm$ 0.36 & \cellcolor[HTML]{C0C0C0}1.2508 $\pm$ 0.23 $\approx$ \\
			SYM-PART          & 7.1278 $\pm$ 2.36 & \cellcolor[HTML]{C0C0C0}0.0647 $\pm$ 0.00 $+$ &  & 6.6313 $\pm$ 2.79 & \cellcolor[HTML]{C0C0C0}2.0840 $\pm$ 2.32 $+$       \\
			MMF1              & 0.1048 $\pm$ 0.03 & \cellcolor[HTML]{C0C0C0}0.0683 $\pm$ 0.00 $+$ &  & 0.1015 $\pm$ 0.02 & \cellcolor[HTML]{C0C0C0}0.0639 $\pm$ 0.00 $+$       \\
			MMF2              & 0.0565 $\pm$ 0.04 & \cellcolor[HTML]{C0C0C0}0.0205 $\pm$ 0.01 $+$ &  & 0.0734 $\pm$ 0.04 & \cellcolor[HTML]{C0C0C0}0.0422 $\pm$ 0.03 $+$       \\
			MMF3              & 0.0413 $\pm$ 0.02 & \cellcolor[HTML]{C0C0C0}0.0181 $\pm$ 0.01 $+$ &  & 0.0386 $\pm$ 0.02 & \cellcolor[HTML]{C0C0C0}0.0292 $\pm$ 0.03 $+$       \\
			MMF4              & 0.1659 $\pm$ 0.07 & \cellcolor[HTML]{C0C0C0}0.0432 $\pm$ 0.00 $+$ &  & 0.1312 $\pm$ 0.04 & \cellcolor[HTML]{C0C0C0}0.0426 $\pm$ 0.00 $+$       \\
			MMF5              & 0.2003 $\pm$ 0.04 & \cellcolor[HTML]{C0C0C0}0.1095 $\pm$ 0.00 $+$ &  & 0.1843 $\pm$ 0.03 & \cellcolor[HTML]{C0C0C0}0.1054 $\pm$ 0.00 $+$       \\
			MMF6              & 0.2468 $\pm$ 0.06 & \cellcolor[HTML]{C0C0C0}0.0972 $\pm$ 0.00 $+$ &  & 0.1995 $\pm$ 0.04 & \cellcolor[HTML]{C0C0C0}0.1516 $\pm$ 0.02 $+$       \\
			MMF7              & 0.0680 $\pm$ 0.03 & \cellcolor[HTML]{C0C0C0}0.0463 $\pm$ 0.00 $+$ &  & 0.0751 $\pm$ 0.03 & \cellcolor[HTML]{C0C0C0}0.0320 $\pm$ 0.00 $+$       \\
			MMF8              & 1.6918 $\pm$ 0.65 & \cellcolor[HTML]{C0C0C0}0.0999 $\pm$ 0.02 $+$ &  & 1.3505 $\pm$ 0.67 & \cellcolor[HTML]{C0C0C0}0.3508 $\pm$ 0.13 $+$       \\ \midrule
			$+$/$-$/$\approx$ & baseline          & 10/0/0                                        &  & baseline          & 9/0/1                                               \\ \bottomrule\end{tabular}
	}
\end{table}

\begin{table}
	\caption{Statistical comparison results regarding the $\operatorname{IGD}^+$ indicator. Mean and standard deviation of $\operatorname{IGD}^+$ values are shown. Better results are highlighted.}
	\label{table: IGD+ comparison}
	\centering
	\centerline{
		\begin{tabular}{@{\hspace{.5em}}c@{\hspace{2em}}c@{\hspace{1em}}cc@{\hspace{2em}}c@{\hspace{1em}}c@{\hspace{.5em}}}
			\toprule
			                  & NSGA-II                                   & Niching-NSGA-II       &  & SPEA2                                     & Niching-SPEA2         \\ \midrule
			Omni-test         & \cellcolor[HTML]{C0C0C0}0.0100 $\pm$ 0.00 & 0.0192 $\pm$ 0.00 $-$ &  & \cellcolor[HTML]{C0C0C0}0.0081 $\pm$ 0.00 & 0.0138 $\pm$ 0.00 $-$ \\
			SYM-PART          & \cellcolor[HTML]{C0C0C0}0.0081 $\pm$ 0.00 & 0.0124 $\pm$ 0.00 $-$ &  & \cellcolor[HTML]{C0C0C0}0.0074 $\pm$ 0.00 & 0.0138 $\pm$ 0.00 $-$ \\
			MMF1              & \cellcolor[HTML]{C0C0C0}0.0033 $\pm$ 0.00 & 0.0039 $\pm$ 0.00 $-$ &  & \cellcolor[HTML]{C0C0C0}0.0027 $\pm$ 0.00 & 0.0031 $\pm$ 0.00 $-$ \\
			MMF2              & \cellcolor[HTML]{C0C0C0}0.0034 $\pm$ 0.00 & 0.0041 $\pm$ 0.00 $-$ &  & \cellcolor[HTML]{C0C0C0}0.0031 $\pm$ 0.00 & 0.0051 $\pm$ 0.00 $-$ \\
			MMF3              & \cellcolor[HTML]{C0C0C0}0.0032 $\pm$ 0.00 & 0.0039 $\pm$ 0.00 $-$ &  & \cellcolor[HTML]{C0C0C0}0.0029 $\pm$ 0.00 & 0.0034 $\pm$ 0.00 $-$ \\
			MMF4              & \cellcolor[HTML]{C0C0C0}0.0031 $\pm$ 0.00 & 0.0045 $\pm$ 0.00 $-$ &  & \cellcolor[HTML]{C0C0C0}0.0026 $\pm$ 0.00 & 0.0038 $\pm$ 0.00 $-$ \\
			MMF5              & \cellcolor[HTML]{C0C0C0}0.0033 $\pm$ 0.00 & 0.0044 $\pm$ 0.00 $-$ &  & \cellcolor[HTML]{C0C0C0}0.0027 $\pm$ 0.00 & 0.0039 $\pm$ 0.00 $-$ \\
			MMF6              & \cellcolor[HTML]{C0C0C0}0.0033 $\pm$ 0.00 & 0.0043 $\pm$ 0.00 $-$ &  & \cellcolor[HTML]{C0C0C0}0.0027 $\pm$ 0.00 & 0.0039 $\pm$ 0.00 $-$ \\
			MMF7              & \cellcolor[HTML]{C0C0C0}0.0034 $\pm$ 0.00 & 0.0071 $\pm$ 0.00 $-$ &  & \cellcolor[HTML]{C0C0C0}0.0029 $\pm$ 0.00 & 0.0034 $\pm$ 0.00 $-$ \\
			MMF8              & \cellcolor[HTML]{C0C0C0}0.0026 $\pm$ 0.00 & 0.0034 $\pm$ 0.00 $-$ &  & \cellcolor[HTML]{C0C0C0}0.0023 $\pm$ 0.00 & 0.0034 $\pm$ 0.00 $-$ \\ \midrule
			$+$/$-$/$\approx$ & baseline                                  & 0/10/0                &  & baseline                                  & 0/10/0                \\ \bottomrule
		\end{tabular}}
\end{table}

Table \ref{table: IGDX comparison} and Table \ref{table: IGD+ comparison} present the statistical comparison results regarding the IGDX and $\operatorname{IGD}^+$ indicators, respectively. In each table, the Wilcoxon rank-sum test is performed with $p=0.05$ to compare the performance of Niching-SPEA2 and Niching-NSGA-II with their original algorithms. The symbols "$+$", "$-$", and "$\approx$" in each table indicated that the corresponding algorithm is outperform, underperform, and tied with the baseline in the statistical comparison.

Table \ref{table: IGDX comparison} clearly shows the superiority of Niching-SPEA2 and Niching-NSGA-II in comparison to the original versions regarding the IGDX indicator. That is, the two modified algorithms have significantly smaller IGDX values than the corresponding original algorithms for almost all test problems. The statistical comparison results further verify that the proposed niching approach can significantly improve the performance of SPEA2 and NSGA-II on various MMOPs. In Table \ref{table: IGD+ comparison}, we can see that the modified algorithms with the niching strategy have worse $\operatorname{IGD}^+$ values on all test problems than the original ones. This is consistent with the our previous observations on Fig. \ref{fig: NSGA-II on SYM-PART} and Fig. \ref{fig: Niching-NSGA-II on SYM-PART} (i.e., there exists a trade-off between the diversity in the decision and the objective spaces). However, from careful examinations of Table \ref{table: IGDX comparison} and Table \ref{table: IGD+ comparison}, we can see for many test problems that large improvement of the IGDX values in Table \ref{table: IGDX comparison} is obtained at the cost of small deterioration of the $\operatorname{IGD}^+$ values in Table \ref{table: IGD+ comparison}. The observation above demonstrates that the proposed niching strategy is a promising approach to multi-modal multi-objective optimization problem, whereas the handling of the trade-off remains an open question.
\section{Concluding Remarks}
\label{sec: Conclusion}
In this paper, we proposed a niching diversity estimation method for multi-modal multi-objective optimization. First, we pointed out that standard diversity estimators in MOEAs meet some challenges when handling MMOPs. To address this issue, we proposed a general niching strategy which is applicable to existing MOEAs to enhance their performance on MMOPs. In our proposed niching strategy, only neighboring solutions in the decision space are involved in diversity estimation. In this manner, the proposed niching strategy is able to prevent the loss of equivalent Pareto optimal solutions. In our experimental studies, we incorporated the proposed niching strategy into two classical MOEAs: SPEA2 and NSGA-II. Experimental results on ten MMOPs clearly showed that the performance of the modified algorithms is notably improved compared to the original algorithms. Currently, we employed a simple niching strategy based on the $k$-th nearest neighbor. The value of $k$ was also simply specified by the square root of the sample size without considering the dimensionality of the decision space. The major contribution of this paper was to clearly illustrate that the incorporation of such a simple niching strategy significantly improved the performance of existing MOEAs on MMOPs. A future research direction can be examining the effect of the value of $k$ and to propose a more effective specification method. 
More experiments over various test problems using a wide variety of MOEAs can be conducted to examine the 
effects of our proposed niching mechanism on MOEAs. Moreover, another promising future research issue is developing more sophisticated and efficient niching strategies. In this research direction, the point may be how to handle the trade-off between the decision space performance and the objective space performance.

\section*{Acknowledgements}
This work was supported by National Natural Science Foundation of China (Grant No. 61876075), Guangdong Provincial Key Laboratory Grant(No. 2020B121201001), the Program for Guangdong Introducing Innovative and Enterpreneurial Teams (Grant No. 2017ZT07X386), Shenzhen Science and Technology Program (Grant No. KQTD2016112514355531), the Program for University Key Laboratory of Guangdong Province (Grant No. 2017KSYS008).
%
%
\bibliographystyle{splncs04}
\bibliography{reference.bib}
\end{document}